%% file: main_arxiv.tex
\documentclass{article} 
\usepackage{iclr2022_conference,times}

\input{math_commands.tex}

\input{tex/00_abbrev_and_macro}

\usepackage{hyperref}
\usepackage{url}

\usepackage{adjustbox}
\usepackage{caption}
\usepackage{subcaption}
\usepackage{booktabs}

\title{Large-scale Bilingual Language-Image Contrastive Learning}

\author{Byungsoo Ko\thanks{Authors contributed equally.},\space\space Geonmo Gu\footnotemark[1]\\
NAVER Vision\\
{\tt\small \{kobiso62, korgm403\}@gmail.com}
}

\iclrfinalcopy 
\begin{document}

\maketitle

\input{tex/00_abstract}
\input{tex/01_introduction}
\input{tex/02_related}
\input{tex/03_dataset}
\input{tex/04_method}

\input{tex/05_experiments}

\input{tex/06_conclusion}

\bibliography{iclr2022_conference}
\bibliographystyle{iclr2022_conference}

\appendix
\input{tex/99_appendix}

\end{document}

%% file: math_commands.tex
\usepackage{amsmath,amsfonts,bm}

\def\eqref#1{equation~\ref{#1}}

\def\1{\bm{1}}

\DeclareMathAlphabet{\mathsfit}{\encodingdefault}{\sfdefault}{m}{sl}
\SetMathAlphabet{\mathsfit}{bold}{\encodingdefault}{\sfdefault}{bx}{n}



%% file: tex/00_abbrev_and_macro.tex
\usepackage{xcolor}
\usepackage{xspace}
\usepackage{diagbox}

\usepackage[utf8]{inputenc}
\usepackage{kotex}
\usepackage{adjustbox}
\usepackage{wrapfig}
\usepackage{multirow}
\newcommand*\rot{\rotatebox{90}}

\makeatletter
\DeclareRobustCommand\onedot{\futurelet\@let@token\@onedot}
\def\@onedot{\ifx\@let@token.\else.\null\fi\xspace}

\def\ie{\emph{i.e}\onedot}

\makeatother

\newcommand{\cL}{\mathcal{L}}

%% file: tex/00_abstract.tex
\begin{abstract}

This paper is a technical report to share our experience and findings building a Korean and English bilingual multimodal model. While many of the multimodal datasets focus on English and multilingual multimodal research uses machine-translated texts, employing such machine-translated texts is limited to describing unique expressions, cultural information, and proper noun in languages other than English. In this work, we collect 1.1 billion image-text pairs (708 million Korean and 476 million English) and train a bilingual multimodal model named KELIP. We introduce simple yet effective training schemes, including MAE pre-training and multi-crop augmentation. Extensive experiments demonstrate that a model trained with such training schemes shows competitive performance in both languages. Moreover, we discuss multimodal-related research questions: 1) strong augmentation-based methods can distract the model from learning proper multimodal relations; 2) training multimodal model without cross-lingual relation can learn the relation via visual semantics; 3) our bilingual KELIP can capture cultural differences of visual semantics for the same meaning of words; 4) a large-scale multimodal model can be used for multimodal feature analogy. We hope that this work will provide helpful experience and findings for future research. We provide an open-source pre-trained KELIP~\footnote{\url{https://github.com/navervision/KELIP}}.

\end{abstract}

%% file: tex/01_introduction.tex
\section{Introduction}
\label{sec:introduction}

Recent years have witnessed the success of large-scale vision and language pre-training models~\citep{yao2021filip,jain2021mural,shonenkov2022ruclip,mu2021slip,li2022blip}, such as CLIP~\citep{radford2021learning} and ALIGN~\citep{jia2021scaling}. Those multimodal models learn visual and textual representations by exploiting millions of image-text pairs collected from the Internet and show state-of-the-art performance on various downstream tasks. In order to learn visual and textual representations, the entire framework is composed of two separated models by each modality and trained to embed semantically similar visual and textual representations to be close and dissimilar representations to be far apart. Such a modality-wise separated model has several advantages. As getting a large number of image-text pairs is costly, each unimodal model can be pre-trained separately. Moreover, each unimodal model can be used separately for different types of downstream tasks, such as text-driven image manipulation~\citep{patashnik2021styleclip}, text to image generation~\citep{galatolo2021generating}, image captioning~\citep{mokady2021clipcap}, video-text retrieval~\citep{luo2021clip4clip}, and large classes detection~\citep{zhou2022detecting}.

Despite the powerful performance and generalization ability to various multimodal tasks, those models are not easy to use for different languages because the models are trained mostly in English, and there can be bias to the cultures and regions of the primary language of the dataset~\citep{fabbrizzi2021survey,mehrabi2021survey}. There have been studies to embed multilingual representation to multimodal models~\citep{jain2021mural} using machine-translated texts. However, such biases still remain because visual biases do not change although the language changes. For example, for the word `breakfast', a picture of western breakfast can be with eggs and bread with bacon, while a picture of Asian breakfast can be rice and soup. Because of such biases and limitation of describing unique expressions, cultural information, and proper noun in languages other than English, the published model can be limited in use to other language-speaking areas.

In this paper, we present a bilingual multimodal model, named KELIP for Korean and English bilingual contrastive Language-Image Pre-training. We use 1.1 billion image-text pairs (708 million Korean and 476 million English) to train KELIP, where every text is not machine translated. We introduce simple but effective training schemes to train large-scale multimodal models. KELIP successfully expert to both Korean and English, which is proven by extensive experiments that shows competitive performance for both languages in quantitative and qualitative results.

Moreover, we discuss research questions related to multimodal models: 1) strong visual augmentation (\ie color jittering) can distract the model from learning the correct relation between visual and text semantics, 2) a model trained without cross-lingual relation can learn the relation of different languages via visual semantics, 3) our bilingual KELIP can capture different characteristics of each cultural vision information for the same meaning of words but different languages, and 4) a large-scale vision and language model can perform simple multimodal feature analogy (\ie an image of field + `a house' in text = an image of a house on a field). Note that this paper is a technical report to share our experience and findings when building a bilingual multimodal model and discuss related research questions.

%% file: tex/02_related.tex
\section{Related Work}
\label{sec:related}

The pre-training and then fine-tuning technique has shown remarkable achievement in natural language processing~\citep{radford2018improving,devlin2019bert,brown2020language} and computer vision~\citep{chen2020simclr,he2020moco,grill2020bootstrap,caron2020swav,he2021masked}. Recently, such technique has been extended to vision and language pre-training, which can be categorized into two: 1) based on vision and language unified model, UNITER~\citep{chen2020uniter}, VisualBERT~\citep{li2019visualbert}, ViLBERT~\citep{lu2019vilbert} and DALLE~\citep{ramesh2021zero} exploit Language Modeling (LM) objectives, such as masked LM and autoregressive LM; 2) based on separated vision and language models, CLIP~\citep{radford2021clip} and ALIGN~\citep{jia2021scaling} employ cross-modal contrastive learning which aims to align visual and textual information on the unified embedding space. In this paper, we will focus on the second category of language-image contrastive learning.

When CLIP was published, it quickly received great attention for its architectural simplicity, large scale, and powerful performance. CLIP shows that, with large-scale datasets (400 million image and text pairs), simple language-image contrastive learning can achieve superior zero-shot ability and robustness. As public image-text pair datasets (\ie YFCC100M~\citep{thomee2016yfcc100m}, CC12M~\citep{sharma2018conceptual}) are involved in the non-trivial cleaning process, ALIGN~\citep{jia2021scaling} exploits larger yet uncurated datasets, and MURAL~\citep{jain2021mural} extends ALIGN to multilingual by presenting cross-lingual objectives. BASIC~\citep{pham2021combined} scales up the contrastive learning framework of CLIP and ALIGN in terms of data size, model size, and batch size to achieve better zero-shot performance. SLIP~\citep{mu2021slip} and DeCLIP~\citep{li2021supervision} employ self-supervised objectives to enhance the individual encoder's representation. In this paper, while following the dual-stream architecture for simplicity and efficiency as CLIP, we further train our model with a large-scale bilingual dataset and modified learning schemes.

%% file: tex/03_dataset.tex
\section{Dataset}
\label{sec:dataset}

This section presents the details of the dataset we used for KELIP, including what kind of public dataset we used and how we gathered additional datasets. We use about 1.1 billion datasets in total, which includes 476 million of English datasets (Sec.~\ref{sec:english}) and 708 million of Korean datasets (Sec.~\ref{sec:korean}).

\subsection{English Dataset}
\label{sec:english}

We collect possible public datasets of multimodal pairs (image and English text). First of all, CUB200~\citep{wah2011cub} is a small dataset of bird images with ten single-sentence descriptions for each image. Wikipedia-based Image Text dataset (WIT)~\citep{srinivasan2021wit} is a large-scale multimodal and multilingual dataset, which is composed set of 37.4 million image-text samples with 11.5 million unique images from 108 languages. The dataset includes images and their contextual metadata from Wikipedia. YFCC15M is a subset of YFCC100M~\citep{thomee2016yfcc100m}, which is filtered in CLIP to keep only images with natural language titles and/or descriptions in English. Moreover, we use the Conceptual Captions (CC3M)~\citep{sharma2018conceptual} and the Conceptual 12M (CC12M)~\citep{sharma2018conceptual}, which are another large scale image-text pairs of datasets.

LAION400M~\citep{schuhmann2021laion} dataset is a 400M of large-scale image-text pair dataset extracted from the Common Crawl (CC)~\footnote{https://commoncrawl.org} web data dump. Following the data construction process of LAION400M, we further created a new dataset of 70 million (image and English text) pairs collected from the CC web data dump. First, we extracted a pair of image URLs and English alt texts from the CC data dump. Second, We downloaded images which is larger than 32 pixels for the smaller side and saved them as 256 $\times$ 256 size. Third, we used the CLIP ViT-B/32 model to compute the similarity between images and texts; then we removed pairs with the similarity below 0.28. At the same time, we computed the similarity between images and predefined NSFW texts; then we removed pairs with a similarity larger than 0.22 to discard NSFW contents. For the total English datasets, we could collect about 476 image-text pairs, excluding images that were not downloaded.

\subsection{Korean Dataset}
\label{sec:korean}

As multimodal research has been mainly focused on English, most of the publicly available datasets are in English, and there is a lack of other language datasets. Many of the publicly available multilingual datasets or multilingual research adopt machine-translated datasets. However, such machine-translated datasets have limitations in describing unique expressions, cultural information, proper noun, and so on. With such motivation, we constructed a new Korean dataset of 708 million. The majority of the dataset is collected from a variety of publicly available sources, especially Korean websites on the Internet. We included 50 million of the celebrities' photos and names to make KELIP have face recognition ability. Moreover, Korean Wikipedia datasets are included. As a result, 708 million of image and Korean text pairs are collected, and the resulting dataset is larger than the LAION400M or total datasets used in CLIP.

%% file: tex/04_method.tex
\section{Method}
\label{sec:method}

In this section, we present our training schemes used in KELIP. In phase 1, we first pre-train the image encoder in a self-supervised manner (Sec.~\ref{sec:pre-training}). In phase 2, we fine-tune the multimodal model in a contrastive learning manner with the multi-crop~\cite{caron2020swav} technique (Sec.~\ref{sec:multi-modal}).

\begin{figure}[t]
    \centering
    \begin{subfigure}[c]{0.52\textwidth}
        \centering
        \includegraphics[width=\textwidth]{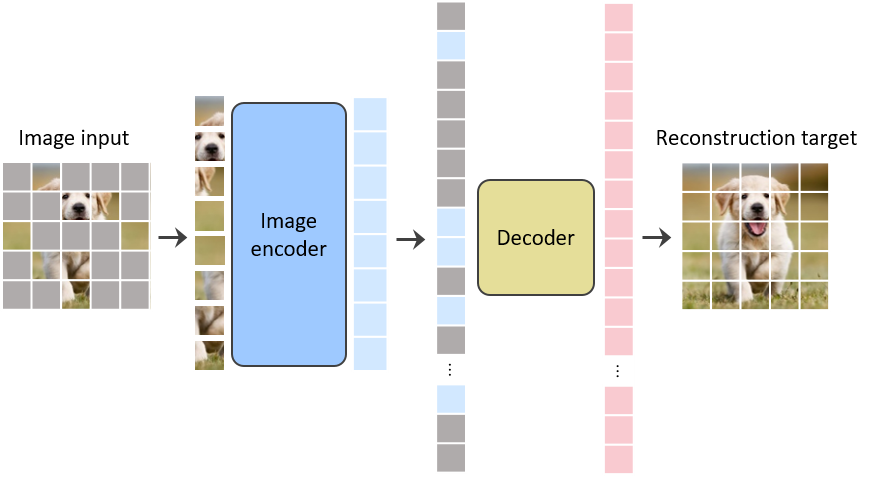}
        \caption{Phase 1: MAE pre-training}
        \label{fig:mae}
    \end{subfigure}
    \begin{subfigure}[c]{0.46\textwidth}
        \centering
        \includegraphics[width=\textwidth]{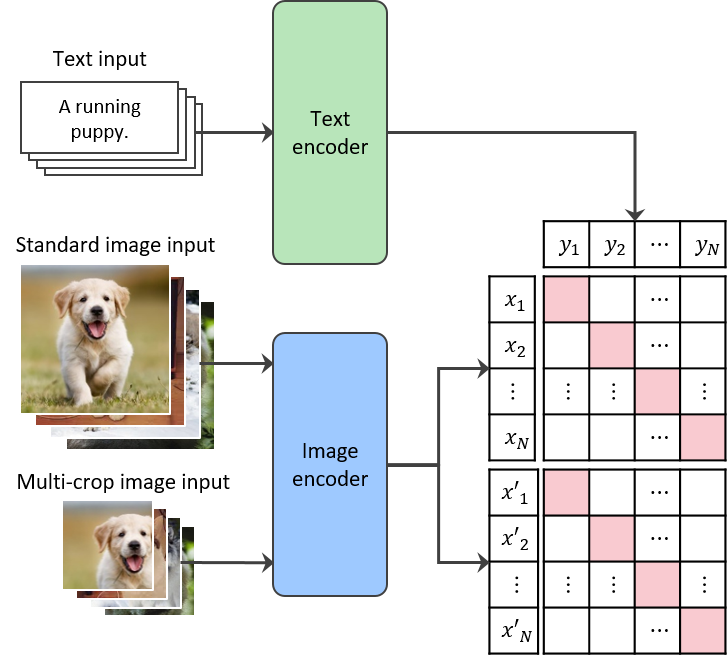}
        \caption{Phase 2: Multi-modal fine-tuning}
        \label{fig:kelip}
    \end{subfigure}%
\caption{\textbf{Summary of our approach}. In phase 1, we pre-train image encoder with MAE methods. Then, we fine-tune multi-modal model based on the pre-trained image encoder in phase 2.}
\label{fig:framework}
\end{figure}

\subsection{Self-supervised pre-training} 
\label{sec:pre-training}

The pre-training and then fine-tuning technique has been proven to be powerful compared to training from scratch~\citep{chen2020simclr,he2020moco,grill2020bootstrap,caron2020swav}. Especially when we train a multimodal model, each modality encoder has limited opportunities to learn with its own modality supervision. Moreover, it has been studied that the performance of the image encoder itself is important when training multimodal model~\citep{zhai2021lit}. With this spirit, we first pre-train image encoder in a self-supervised manner. Self-supervised methods using contrastive learning~\citep{chen2020simclr,he2020moco,caron2020swav} strongly depend on data augmentation and require twice the batch size to get augmented positive pairs, resulting in the slow training process. Instead, we employ the self-supervised method of Masked Autoencoder (MAE)~\citep{he2021masked} with our collected 1.1 billion images, which has been shown fast training and powerful performance. As illustrated in Figure~\ref{fig:mae}, during pre-training, the encoder masks out a large random subset of image patches (\ie 75\%), and then the small decoder reconstructs the original image in pixels. After the pre-training, the decoder is discarded, and the encoder is employed for the multimodal training.

\subsection{Multi-modal training}
\label{sec:multi-modal}

\subsubsection{Model Architecture}
\label{sec:architecture}

Following CLIP, we use a dual-encoder architecture that consists of an image encoder and a text encoder. There is multimodal interaction at the top of the model, where the text and image features are projected to the same size of dimension with L2 normalization. We use a Transformer~\citep{Vaswani2017AttentionIA} with architecture modification~\citep{radford2019language} as text encoder. The text encoder has 63 million parameters, 12 layers, 512 wide model with 8 attention heads. In order to tokenize English or Korean text, we trained a lower-cased byte pair encoding (BPE) representation of the text with 2 million English and 1.6 million Korean texts, which are a subsample of our collected dataset. The resulting BPE contains 98,816 vocab size, which is twice larger than the 49,152 vocab size of CLIP. The max sequence length is limited to 76 for computational efficiency. Every text sequence is bracketed with [SOS] and [EOS] tokens, while the remaining tokens after [EOS] are filled with [PAD]. We use the feature of the last layer in the transformer at the [EOS] as the final text feature, followed by layer normalization and linear projection into the multimodal feature space. For the vision encoder, we use the vision transformer~\citep{dosovitskiy2021an} family (\ie ViT-B/32). We use the feature of the last layer in the ViT at the class token as the final image feature, followed by layer normalization and linear projection into the multimodal feature space.

\subsubsection{Contrastive Learning}
\label{sec:contrastive}

The contrastive objective pushes the matching image-text features together while pulling the unmatched image-text features apart. In a batch of $N$ image-text feature pairs {($x_i$, $y_i$)}, we denote $x_i$ and $y_i$ as image and text features of $i$th pair. Then, we can formulate InfoNCE loss~\citep{oord2018cpc} for image-to-text contrastive learning as follows:
\begin{equation}
    \cL_{I2T} = -\frac{1}{N}\sum_{i=1}^N\log\frac{\text{exp}(\text{sim}(x_i,y_i))/\tau}{\sum_{j=1}^N\text{exp}(\text{sim}(x_i,y_j))/\tau}, 
    \label{eq:infonce}
\end{equation}
where sim($x$,$y$)$=\frac{x^Ty}{\parallel{x}\parallel\parallel{y}\parallel}$ is cosine similarity function computed by dot product of L2 normalized features, and $\tau$ is a learnable temperature to scale the logits. We use the symmetrical loss for text-to-image contrastive learning as $\cL_{T2I}$. The overall loss function for contrastive learning $\cL_{cont}$ is the average of $\cL_{I2T}$ and $\cL_{T2I}$.

\subsubsection{Multi-crop Augmentation}
\label{sec:multi-crop}

Muti-crop augmentation~\citep{caron2020swav} has been proposed in vision self-supervised learning and has shown powerful performance in many works~\citep{caron2021dino,zhou2021ibot}. As mentioned in prior works~\citep{chen2020simclr,misra2020pretext}, comparing the larger number of crops of an image is important to capture information in terms of relations between object and background. However, increasing the number of views can increase the training computation and memory. Thus, multi-crop augmentation employs additional views of low resolution crops that cover small parts of the image, which can ensure a small increase in training computation and memory. We observe that multi-crop augmentation can be used in multimodal training to enhance the final performance. As illustrated in Figure~\ref{fig:kelip}, we extract features of an image with standard resolution (\ie 224), and low resolution (\ie 96) cropped as small parts of the image. Then those increased number of images are computed with corresponding texts with $\cL_{cont}$. As the resolution for multi-crop augmentation is low, the increase in training computation and memory is small, but the model benefits from a performance boost.

%% file: tex/05_experiments.tex
\section{Experiments}
\label{sec:experiment}

In this section, we first share the experimental setting in Sec.~\ref{sec:implementation}, and we show the impact of each component by ablation study in Sec.~\ref{sec:ablation}. Then, we evaluate our KELIP with public benchmarks in Sec.~\ref{sec:quantitative} and show qualitative results in Sec.~\ref{sec:qualitative}. Finally, we discuss multimodal related research questions in Sec.~\ref{sec:discussion}.

\subsection{Experimental Setting}
\label{sec:implementation}
\input{table/tab_ablation}
\input{table/tab_classification}
\input{table/tab_retrieval}

\begin{figure}[t]
    \centering
    \includegraphics[width=1.0\linewidth]{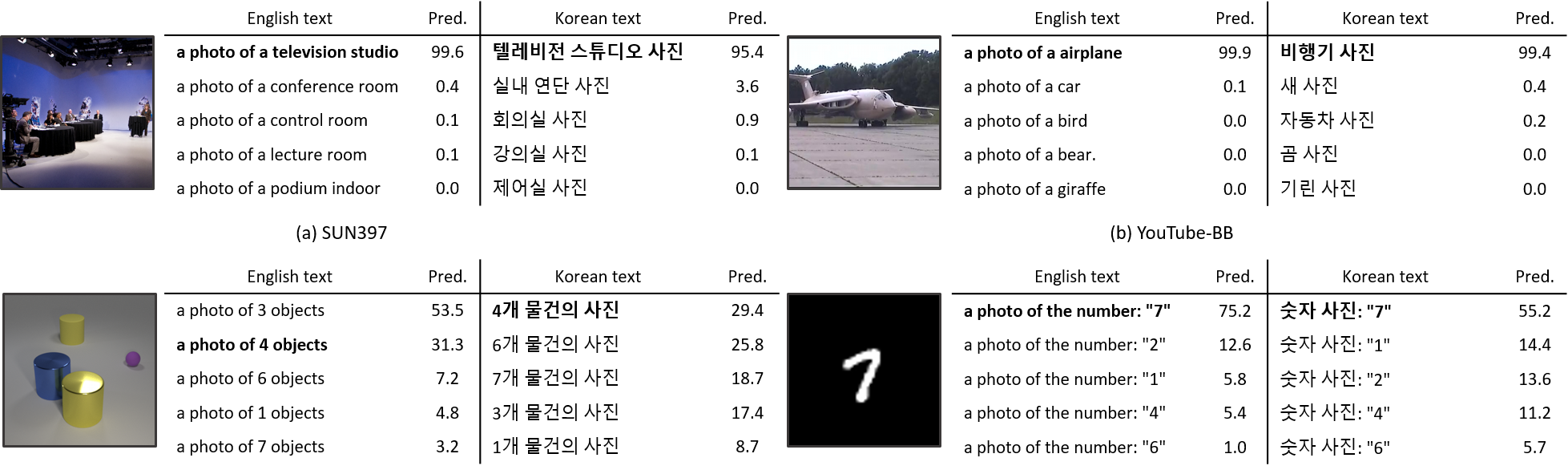}
    \caption{\textbf{Qualitative results.} A zero-shot KELIP classifier classifies each image sample from the public dataset with five English and Korean texts. Bold text indicates ground truth and prediction (Pred.) value is in \%.}
    \label{fig:qualitative}
\end{figure}

\paragraph{Implementation details.} Our implementation is based on PyTorch~\citep{pytorch} framework.
We train ViT-B/32 for KELIP fine-tuning. We use 40 A100 GPUs for KELIP$_{\text{yfcc}}$, which is KELIP trained with YFCC15M, and 80 A100 GPUs for KELIP. MAE pre-training for KELIP took 16 hours, while fine-tuning the multimodal model took 362 hours. For accelerating training time and saving memory, we employ mixed-precision~\citep{micikevicius2017mixed}. The learnable temperature parameter $\tau$ is initialized with 0.07~\citep{wu2018unsupervised}, and it is clipped by 0.01 for training stability. Most of the hyper-parameters of MAE pre-training are followed by~\citep{he2021masked}, while we follow~\citep{radford2021clip} for the hyper-parameters of KELIP fine-tuning. For ablation study, we follow hyper-parameters of SLIP~\citep{mu2021slip}. Different hyper-parameters are chosen by heuristics because of computational constraints. The details of hyper-parameters are shown in Table~\ref{table:hyperparam}.

\paragraph{Benchmark datasets.} For evaluating zero-shot classification, we use the following benchmark datasets: ImageNet~\citep{deng2009imagenet}, Cifar10~\citep{krizhevsky2009learning}, Cifar100~\citep{krizhevsky2009learning}, CLEVR Counts~\citep{johnson2017clevr}, Describable Textures Dataset (DTD)~\citep{cimpoi14describing}, EuroSAT~\citep{helber2019eurosat}, FER2013~\citep{goodfellow2013challenges}, Food101~\citep{bossard2014food}, GTSRB~\citep{stallkamp2012man}, MNIST~\citep{lecun1998mnist}, RESISC45~\citep{cheng2017remote}, StanfordCars~\citep{krause20133d}, STL10~\citep{coates2011analysis}. We evaluate them in Korean by translating their English labels to Korean. WebKorean is our private benchmark dataset, which contains 36,826 web images with 428 Korean labels. For zero-shot cross-modal retrieval, we use the following benchmark datasets: Flickr30k~\citep{plummer2015flickr30k}, MSCOCO (English)~\citep{lin2014microsoft} and MSCOCO (Korean)\footnote{https://aihub.or.kr/keti\_data\_board/visual\_intelligence}.

\subsection{Ablation Study}
\label{sec:ablation}

For an ablation study, we train CLIP with YFCC15M by adding each training scheme and evaluate zero-shot classification and cross-modal retrieval. Every experiment follows KELIP$_{\text{yfcc}}$ hyper-parameters in Table~\ref{table:hyperparam} with different training schemes.
As shown in Table~\ref{table:ablation}, fine-tuning CLIP with MAE pre-trained model achieves better performance in both classification and retrieval. When we add one local view of multi-crop, the performance of both tasks is improved significantly. We will further discuss employing SimCLR method in Sec.~\ref{sec:augmentation}.

\subsection{Quantitative Results}
\label{sec:quantitative}

\subsubsection{Zero-shot Classification}
\label{sec:classification}
We evaluate KELIP with 14 benchmark datasets and compare it with CLIP (ViT-B/32) in a zero-shot classification manner. We use 13 English public datasets for English evaluation while translating their labels to Korean for Korean evaluation. Our Korean private dataset (WebKorean) is also used for Korean evaluation. Note that CLIP can be used in Korean due to the flexibility of the BPE tokenizer. As shown in Table~\ref{table:classification}, KELIP shows 3.3\% higher performance on average for English evaluation compared to CLIP. For Korean evaluation, CLIP performances are extremely low because the model is mainly trained in English, while KELIP shows competitive performance for every dataset.

\subsubsection{Zero-shot Cross-modal Retrieval}
\label{sec:retrieval}

Zero-shot cross-modal retrieval consists of image-to-text retrieval and text-to-image retrieval. We compare our KELIP with existing multimodal models, including Visual N-Grams~\citep{li2017learning}, ImageBert~\citep{qi2020imagebert}, Unicoder-VL~\citep{li2020unicoder}, Uniter~\citep{chen2020uniter}, and CLIP (ViT-B/32)~\citep{radford2021clip}. As shown in Table~\ref{table:retrieval}, KELIP achieves the best performance in both MSCOCO (English) and MSCOCO (Korean), where KELIP achieves competitive performance on Flickr30k. Compared to CLIP, our KELIP shows higher performance in most benchmark datasets, demonstrating that KELIP better understands multimodal relations.

\subsection{Qualitative Results}
\label{sec:qualitative}

We perform a qualitative evaluation on images from four benchmark datasets: SUN396~\citep{xiao2010sun}, YouTube-BB~\citep{real2017youtube}, CLEVR Counts~\citep{johnson2017clevr}, and MNIST~\citep{lecun1998mnist}. Each image is classified among 5 Korean and English candidate texts in a zero-shot manner. As shown in Figure~\ref{fig:qualitative} (a) and (b), both indoor and object images are classified correctly for both languages. Moreover, we observe that KELIP can be used for counting objects and recognizing numbers, as shown in Figure~\ref{fig:qualitative} (c) and (d). We further confirm that KELIP can be used for simple face recognition, as shown in Figure~\ref{fig:face}.

\subsection{Discussions}
\label{sec:discussion}

\subsubsection{Limitations of Strong Augmentation-based Training}
\label{sec:augmentation}

\begin{wrapfigure}[18]{R}{0.5\linewidth}
  \vspace{-5pt}  
  \begin{center}
    \resizebox{\linewidth}{!}{
      \includegraphics[width=0.6\linewidth]{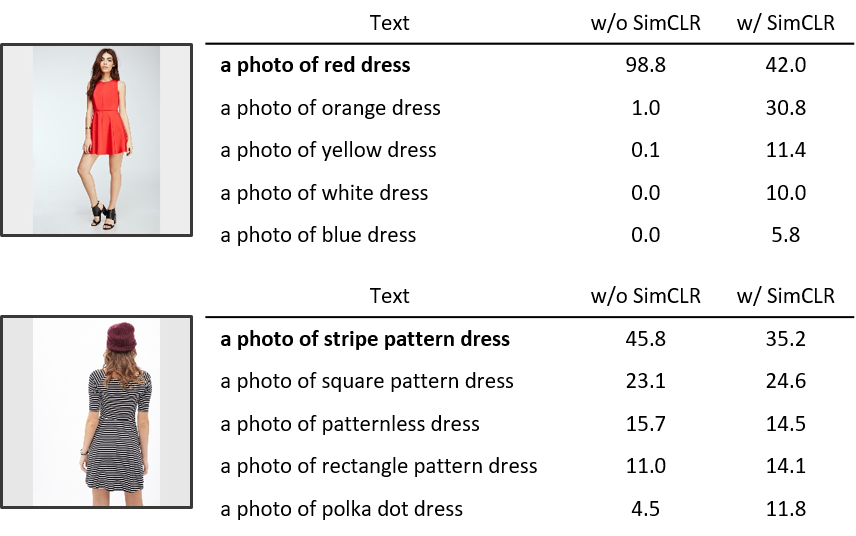}
    }
  \end{center}
  \vspace{-10pt}
\caption{\textbf{Qualitative analysis of employing strong augmentation.} \textit{w/o SimCLR} and \textit{w/ SimCLR} models are \textit{CLIP} and \textit{CLIP + SimCLR} in Table~\ref{table:ablation}, respectively. Prediction scores are in \%.}
\label{fig:augmentation}
\end{wrapfigure}

Recent works (SLIP and DeCLIP) propose to exploit self-supervised learning methods on the image encoder along with multimodal contrastive learning and show competitive performance on zero-shot classification. SLIP and DeCLIP use SimCLR and SimSiam methods, respectively, which contain strong image augmentations (\ie changing brightness, contrast, saturation, hue, etc.). However, we observe that those strong image augmentation can distract the multimodal model from learning the correct relation between modalities. As shown in Figure~\ref{fig:augmentation}, \textit{w/ SimCLR} gets more confused by the color and pattern of the image compared to \textit{w/o SimCLR}. It is also shown in the quantitative evaluation. As shown in Table~\ref{table:ablation}, although the \textit{MAE+CLIP+SimCLR} shows the best performance in zero-shot classification, it shows lower performance than \textit{MAE+CLIP+MultiCrop} in zero-shot cross-lingual retrieval, which requires a higher understanding of cross-lingual relation. Thus, we suggest avoiding strong augmentation for training a multimodal model to expect a profound understanding of cross-lingual relations. With this spirit, we did not employ the self-supervised learning method for multimodal training.

\subsubsection{Cross-lingual Semantic Relation}
\label{sec:cross-lingual}
\begin{figure}[t]
    \centering
    \includegraphics[width=0.75\textwidth]{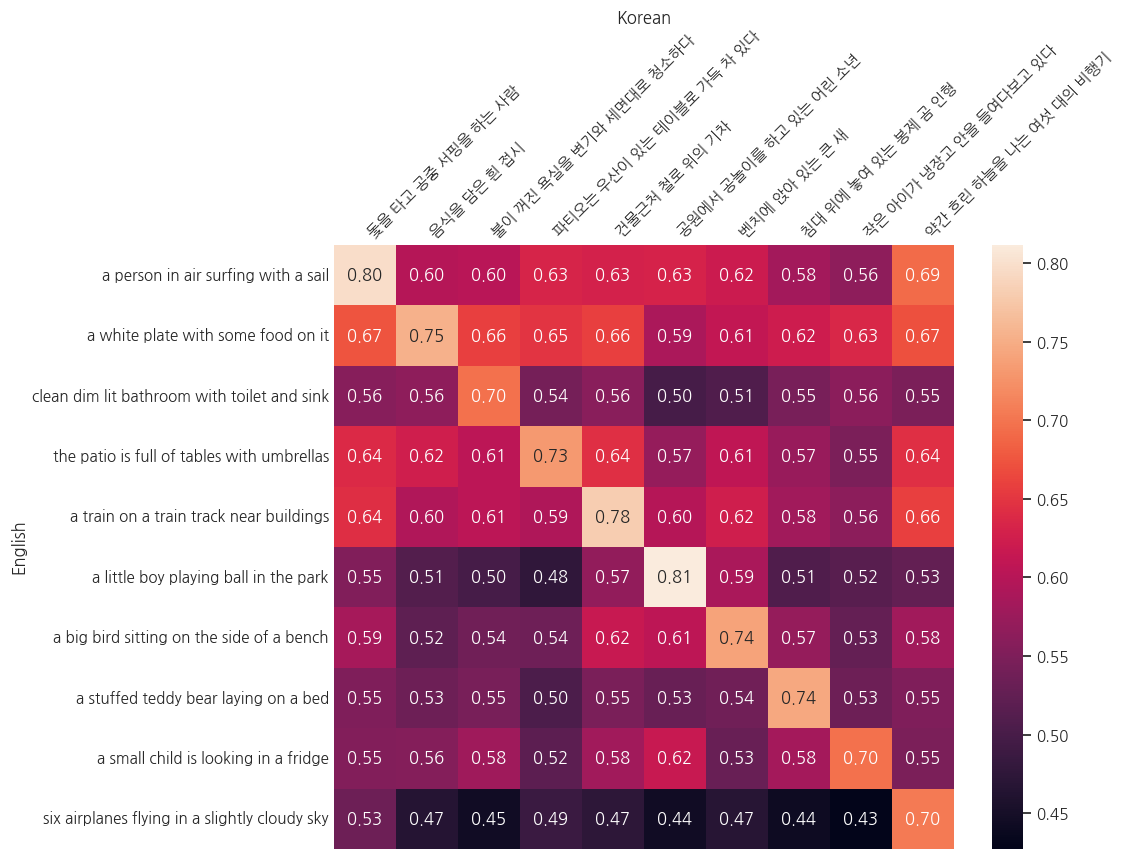}
\caption{\textbf{Cross-lingual cosine similarity heatmap for long sentences}. The same index of each row (English) and column (Korean) has the same semantic meaning in different languages.}
\label{fig:translation}
\end{figure}

When we trained KELIP, we did not put any cross-lingual relations between Korean and English. However, those relations can be learned via shared visual information; \ie the image feature of a photo of an apple would be close to both the English text feature of \textit{apple} and the Korean text feature of \textit{사과}. This can result in an embedding space where semantically similar words in different languages are close and semantically dissimilar words in different languages are far apart. To verify this hypothesis, we conduct an experiment by computing cosine similarities between the same semantic prompts in different languages. As shown in the Figure~\ref{fig:translation} and Figure~\ref{fig:translation_short}, for both cases of long sentences and short words, every sentence gives the highest cosine similarity when the paired sentence has the same semantic. It demonstrates that training KELIP without cross-lingual information can learn the semantic relations between different languages via shared visual information.

\begin{figure}[t]
    \centering
    \includegraphics[width=\linewidth]{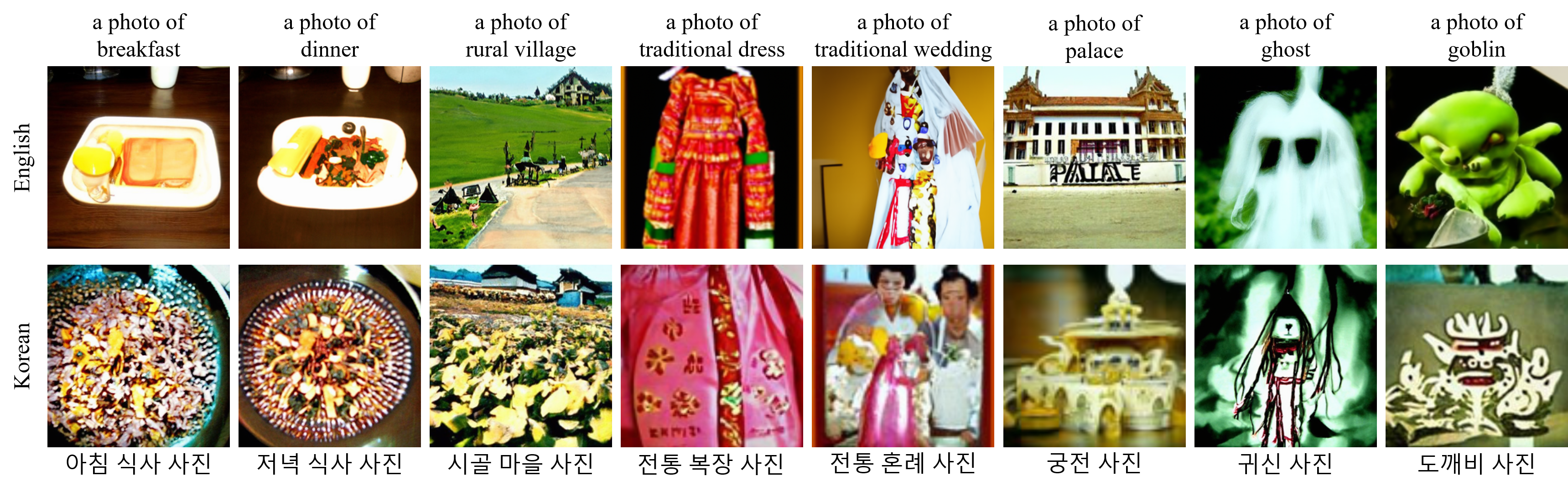}
    \caption{\textbf{KELIP-guided diffusion}, Two images in each column are generated with the same semantic prompt but different languages; English in the first row and Korean in the second row.}
    \label{fig:cgd}
\end{figure}

\subsubsection{How does KELIP sees?}
\label{sec:diffusion}
As language is the way by which people communicate with one another, it reflects cultures. Same semantic words but different languages can have different visual information because of cultural differences. As each language dataset includes its visual information in terms of culture, this raises a question: \textit{how does KELIP see to the same semantic words but different languages?} To answer the question, we conduct an experiment generating images with given text prompts by KELIP-guided diffusion~\citep{dhariwal2021diffusion,ho2020denoising}. With such text-to-image generation, we can generate images of abstracted textual information.

As shown in the Figure~\ref{fig:cgd}, \textit{breakfast} and \textit{dinner} in English look like bread and salad, reflecting general western meal, while those of Korean look like dishes with rice, reflecting general Asian meal. \textit{Traditional dress} and \textit{traditional wedding} in Korean show the visual patterns of Korean traditional clothes called `hanbok (한복)', while those of English show general long and slim dresses. The concepts of \textit{ghost} and \textit{goblin} are also different by culture. The general image of a ghost in the western is floating white cloth with two black holes, while that of an Asian can be a person with long hair and a white dress. For the goblin, the general image in the western is a green monster with large ears, while the goblin in Korea has horns with a thick beard. Those characteristics are depicted in Figure~\ref{fig:cgd}. It demonstrates that the bilingual KELIP model captures different characteristics of each cultural vision information for the same semantic meaning of words but different languages.

\subsubsection{Multimodal Feature Analogy}
\label{sec:analogy}

\begin{figure}[t]
    \centering
    \includegraphics[width=0.8\linewidth]{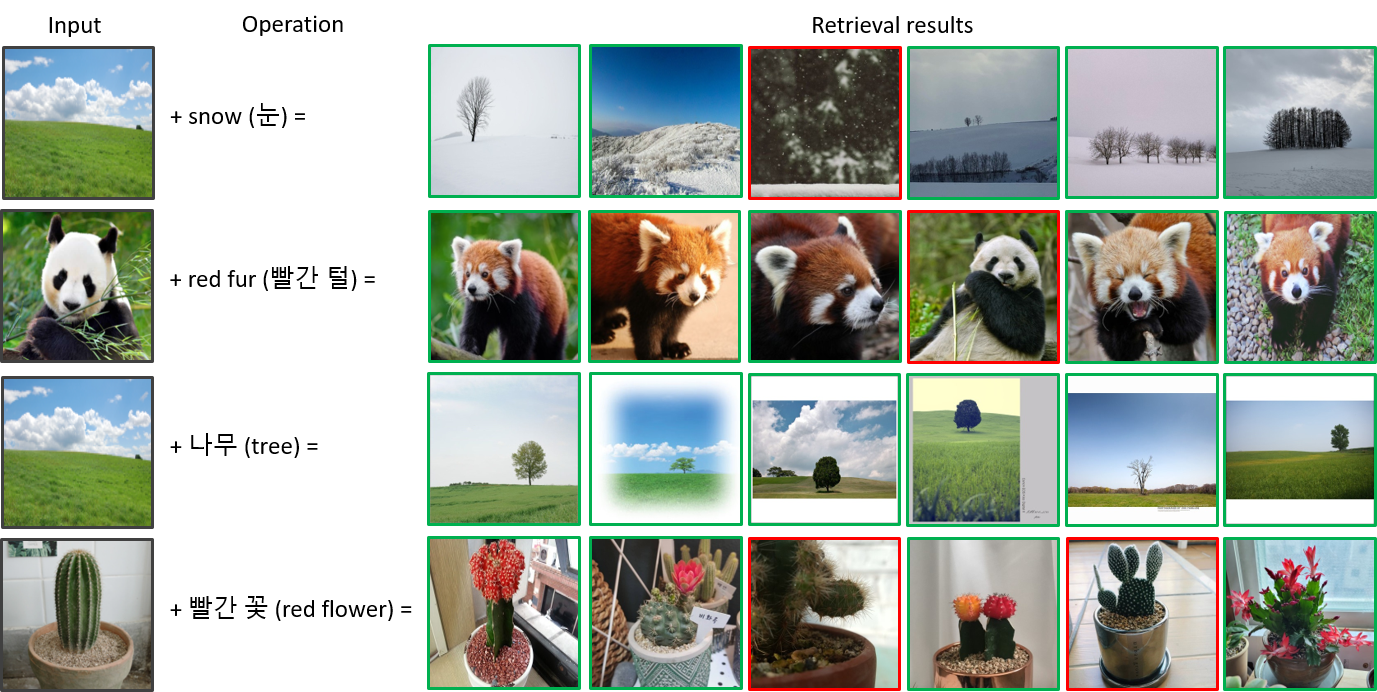}
    \vspace{-2mm}
    \caption{\textbf{Multimodal feature analogy.} We construct query feature by arithmetic operation of image and text feature; then, we perform image retrieval from the gallery set. Parentheses are the translation of the text inputs.}
    \label{fig:analogy}
\end{figure}

There have been demands of studies to combine more than two different modalities in feature analogy~\citep{shin2021rtic,ben2017mutan,fukui2016multimodal,yu2017multi,ethayarajh2018towards}. To see if our KELIP can be used for multimodal feature analogy, we conducted an experiment by performing image retrieval with a query feature that is fused with image and text features. We first construct a gallery set with 99 million of random images. Then, we compute a query feature by $q=l_2(x+wy)$, where $l_2()$ is L2 normalization and we heuristically find the best $w$ within 1 to 5 for the best results. As shown in Figure~\ref{fig:analogy}, every example successfully combines multimodal features and retrieves images that contain semantics of text based on the input images. It demonstrates the potential of our KELIP model to be used for multimodal feature analogy.

%% file: table/tab_ablation.tex
\begin{table}[t]
    \centering
    \begin{subtable}[h]{0.9\textwidth}
        \begin{adjustbox}{width=1.0\textwidth,center}
        \centering
        \begin{tabular}{lcccccccccccccc}
        \toprule
        Method $\backslash$ Benchmark     & \rot{ImageNet} & \rot{Cifar10} & \rot{Cifar100} & \rot{CLEVR-C} & \rot{DTD}  & \rot{EuroSAT} & \rot{FER2013} & \rot{Food101} & \rot{GTSRB} & \rot{MNIST} & \rot{RESISC45} & \rot{StanfordCars} & \rot{STL10} & \rot{Average} \\
        \midrule
        CLIP                   & 33.3     & \textbf{67.7}    & 33.7     & 14.0          & 18.1 & \textbf{31.7}    & 18.4    & 43.1    & 10.3  & 6.9   & 25.4     & 4.5          & 90.4  & 30.6    \\
        MAE+CLIP           & 35.2     & 64.8    & \textbf{34.2}     & \textbf{14.4}          & 17.0 & 21.0    & 22.8    & 44.5    & \textbf{13.4}  & 11.5  & 23.1     & \textbf{4.9}          & 91.8  & 30.7    \\
        MAE+CLIP+MultiCrop & \textbf{38.5}     & 64.9    & 33.5     & 12.0          & \textbf{21.2} & 27.0    & \textbf{24.1}    & \textbf{48.0}    & 10.4  & \textbf{15.7}  & \textbf{27.7}     & 4.7          & \textbf{93.5}  & \textbf{32.4}    \\
        \midrule
        MAE+CLIP+SimCLR    & \textbf{39.9}     & 79.1    & 46.5     & \textbf{14.1}          & 21.0 & 29.2    & \textbf{20.7}    & \textbf{49.3}    & \textbf{7.3}   & \textbf{11.6}  & \textbf{25.1}     & \textbf{7.2}          & 93.5  & \textbf{34.2}    \\
        CLIP+SimCLR            & 39.3     & \textbf{80.5}    & \textbf{47.6}     & 13.2          & \textbf{21.7} & \textbf{30.4}    & 19.4    & 49.1    & 7.1   & 10.0  & 24.4     & 6.9          & \textbf{93.7}  & 34.1   \\
        \bottomrule
        \end{tabular}
        \end{adjustbox}
    \caption{Zero-shot classification}
    \vspace{2mm}
    \label{table:ablation_cls}
    \end{subtable}
    \begin{subtable}[h]{0.9\columnwidth}
        \begin{adjustbox}{width=1.0\columnwidth,center}
        \centering
        \begin{tabular}{lcccccccccccc}
        \toprule
        Task $\rightarrow$               & \multicolumn{6}{c}{Image-to-text-retrieval}                             & \multicolumn{6}{c}{Text-to-image retrieval}                     \\
        Banchmark $\rightarrow$              & \multicolumn{3}{c}{Flickr30k} & \multicolumn{3}{c}{MSCOCO (English)} & \multicolumn{3}{c}{Flickr30k} & \multicolumn{3}{c}{MSCOCO (English)} \\
        Method $\downarrow$                & R@1      & R@5      & R@10    & R@1          & R@5         & R@10       & R@1      & R@5      & R@10    & R@1       & R@5      & R@10     \\
        \midrule
        CLIP                   & 43.9     & 73.2     & 82.6    & 25.1         & 49.1        & 60.1       & 27.3     & 53.7     & 64.7    & 14.8      & 34.4     & 45.9     \\
        MAE+CLIP           & 44.5     & 72.6     & 83.0    & 26.0         & 50.4        & 61.5       & 27.4     & 53.2     & 64.1    & 15.6      & 35.8     & 47.0     \\
        MAE+CLIP+MultiCrop & \textbf{51.7}     & \textbf{77.7}     & \textbf{87.9}    & \textbf{29.9}         & \textbf{54.6}        & \textbf{65.8}       & \textbf{32.7}     & \textbf{59.2}     & \textbf{69.4}    & \textbf{17.9}      & \textbf{39.4}     & \textbf{51.2}     \\
        \midrule
        MAE+CLIP+SimCLR    & \textbf{48.6}     & \textbf{76.0}     & \textbf{86.0}    & \textbf{29.1}         & \textbf{54.8}        & \textbf{66.3}       & \textbf{31.6}     & \textbf{57.7}     & \textbf{68.9}    & 17.5      & \textbf{38.9}     & 51.1     \\
        CLIP+SimCLR            & 48.1     & 75.5     & \textbf{86.0}    & 28.7         & 54.1        & 65.6       & 31.3     & 57.3     & 68.6    & \textbf{17.6}      & 38.8     & \textbf{51.2}     \\
        \bottomrule
        \end{tabular}
        \end{adjustbox}
    \caption{Zero-shot cross-modal retrieval}
    \label{table:ablation_retrieval}
    \end{subtable}
\caption{\textbf{Ablation study.} We use one local view for multi-crop. CLEVR-C indicates CLEVR Counts dataset and all values are in \%.}
\label{table:ablation}
\end{table}

%% file: table/tab_classification.tex
\begin{table}[t]
\begin{adjustbox}{width=0.85\textwidth,center}
\centering
\begin{tabular}{lcccccccccccccccc}
\toprule
                         &       & \rot{ImageNet} & \rot{WebKorean} & \rot{Cifar10} & \rot{Cifar100} & \rot{CLEVR-C} & \rot{DTD}   & \rot{EuroSAT} & \rot{FER2013} & \rot{Food101} & \rot{GTSRB} & \rot{MNIST} & \rot{RESISC45} & \rot{StanfordCars} & \rot{STL10} & \rot{Average} \\
\midrule
\multirow{2}{*}{\rot{Eng.}} & CLIP                 & \textbf{63.4}     & -      & 86.9    & 59.7     & \textbf{20.6}          & 44.4 & 45.3    & \textbf{49.0}    & \textbf{82.3}    & 32.6  & 48.3  & \textbf{60.0}     & 59.6         & \textbf{97.1}  & 57.6    \\
                         & KELIP                & 62.6     & -      & \textbf{91.5}    & \textbf{68.6}     & 13.2          & \textbf{51.2} & \textbf{59.0}    & 37.5    & 79.5    & \textbf{37.6}  & \textbf{60.0}  & 59.6     & \textbf{75.4}         & 96.1  & \textbf{60.9}    \\
\midrule
\multirow{2}{*}{\rot{Kor.}}  & CLIP                 & 0.4      & 2.5      & 15.6    & 2.6      & 11.4          & 1.0  & 11.8    & 13.9    & 1.5     & 15.5  & 3.4   & 9.6      & 16.6         & 16.6  & 8.7     \\
                         & KELIP                & \textbf{38.3}     & \textbf{57.7}     & \textbf{91.4}    & \textbf{55.5}     & \textbf{16.8}          & \textbf{26.9} & \textbf{29.5}    & \textbf{5.4}     & \textbf{46.0}    & \textbf{3.9}   & \textbf{63.4}  & \textbf{42.5}     & \textbf{53.7}         & \textbf{94.2}  & \textbf{44.7}  \\
\bottomrule
\end{tabular}
\end{adjustbox}
\caption{\textbf{Zero-shot classification.} Every benchmark is English public dataset except WebKorean, which is Korean private dataset. ``Kor.'' denotes evaluation with English to Korean translated labels. CLEVR-C indicates CLEVR Counts and values are in Recall (\%).}
\label{table:classification}
\end{table}

%% file: table/tab_retrieval.tex
\begin{table}[t]
\begin{adjustbox}{width=1.0\textwidth,center}
\centering
\begin{tabular}{lcccccccccccccccccc}
\toprule
Task $\rightarrow$          & \multicolumn{9}{c}{Image-to-text retrieval}                                                                & \multicolumn{9}{c}{Text-to-image retrieval}                                                                \\
Benchmark $\rightarrow$      & \multicolumn{3}{c}{Flickr30k} & \multicolumn{3}{c}{MSCOCO (English)} & \multicolumn{3}{c}{MSCOCO (Korean)} & \multicolumn{3}{c}{Flickr30k} & \multicolumn{3}{c}{MSCOCO (English)} & \multicolumn{3}{c}{MSCOCO (Korean)} \\
Method $\downarrow$        & R@1      & R@5      & R@10    & R@1        & R@5        & R@10       & R@1        & R@5        & R@10      & R@1      & R@5      & R@10    & R@1        & R@5        & R@10       & R@1       & R@5        & R@10       \\
\midrule
Visual N-Grams & 15.4     & 35.7     & 45.1    & 8.7        & 23.1       & 33.3       & -          & -          & -         & 8.8      & 21.2     & 29.9    & 5.0        & 14.5       & 21.9       & -         & -          & -          \\
ImageBERT      & -        & -        & -       & 44.0       & 71.2       & 80.4       & -          & -          & -         & -        & -        & -       & 32.3       & 59.0       & 70.2       & -         & -          & -          \\
Unicoder-VL    & 64.3     & 86.8     & 92.3    & -          & -          & -          & -          & -          & -         & 48.4     & 76.0     & 85.2    & -          & -          & -          & -         & -          & -          \\
UNITER         & \textbf{80.7}     & \textbf{95.7}     & \textbf{98.0}    & -          & -          & -          & -          & -          & -         & \textbf{66.2}     & \textbf{88.4}     & \textbf{92.9}    & -          & -          & -          & -         & -          & -          \\
CLIP           & 78.8     & 94.9     & 98.2    & 50.1       & 75.0       & 83.5       & 0.2        & 0.8        & 1.4       & 58.8     & 83.5     & 90.0    & 30.5       & 56.0       & 66.9       & 0.0       & 0.2        & 0.3        \\
KELIP          & 77.4     & 93.9     & 97.2    & \textbf{55.9}       & \textbf{78.7}       & \textbf{86.7}       & \textbf{21.7}       & \textbf{46.3}       & \textbf{59.0}      & 61.6     & 85.7     & 91.5    & \textbf{36.8}       & \textbf{63.3}       & \textbf{73.9}       & \textbf{5.5}       & \textbf{19.0}       & \textbf{28.7}      \\
\bottomrule
\end{tabular}
\end{adjustbox}
\caption{\textbf{Zero-shot cross-modal retrieval.} Flickr20k and MSCOCO (English) are English text based retireval, while MSCOCO (Korean) is Korean text based retrieval. Values are in Recall (\%).}
\label{table:retrieval}
\end{table}

%% file: tex/06_conclusion.tex
\section{Conclusion}
\label{sec:conclusion}

In this paper, we present KELIP, a large-scale bilingual multimodal model, which is trained with 1.1 billion image-text pairs (708 Korean pairs and 476 English pairs) and introduce simple yet effective training schemes for multimodal model. Our ablations show that a model trained with such training schemes can benefit from a large performance boost. Our KELIP model shows competitive performance in both languages. Moreover, we discuss multimodal-related research questions, including strong augmentation-based methods, cross-lingual relation, the cultural difference of textual semantics, and multimodal feature analogy. We left training multimodal model with cross-lingual relation and supporting additional languages as future work.

%% file: tex/99_appendix.tex
\clearpage

\section{Appendix}

\paragraph{Details of hyper-parameter.}
The details of hyper-parameter used in MAE pre-training and multimodal fine-tuning is in Table~\ref{table:hyperparam}.
\input{table/tab_hyperparam}

\begin{figure}[t]
    \centering
    \includegraphics[width=0.6\textwidth]{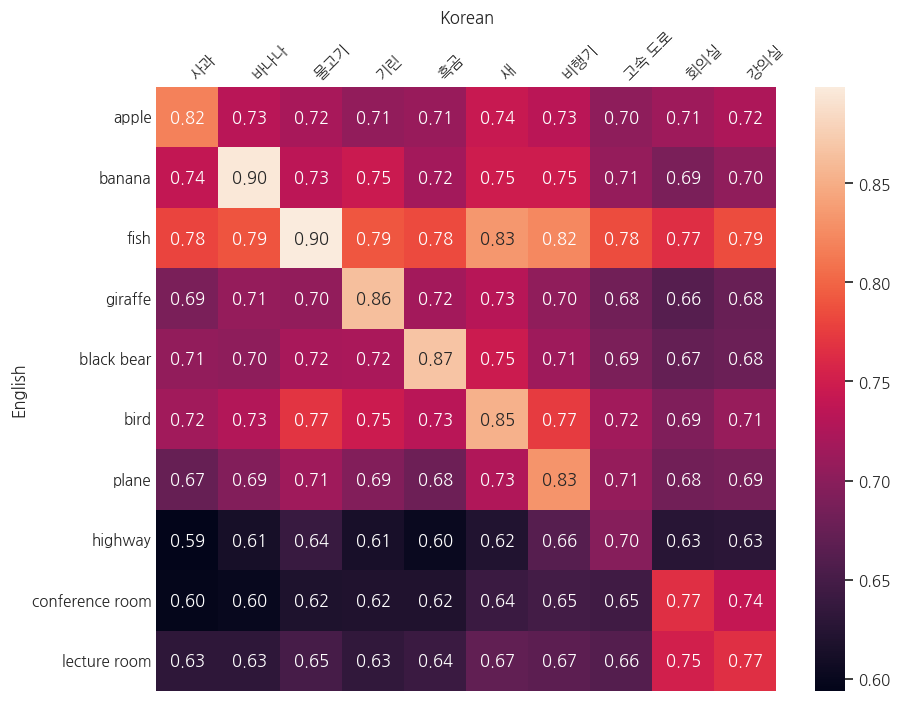}
\caption{\textbf{Cross-lingual cosine similarity heatmap for short words}. The same index of each row (English) and column (Korean) has the same semantic meaning in different languages.}
\label{fig:translation_short}
\end{figure}

\paragraph{Additional Results of Cross-lingual Semantic Relation.}
We include cross-lingual cosine similarity heatmap for short words in Figure~\ref{fig:translation_short}

\begin{figure}[t]
    \centering
    \includegraphics[width=1.0\textwidth]{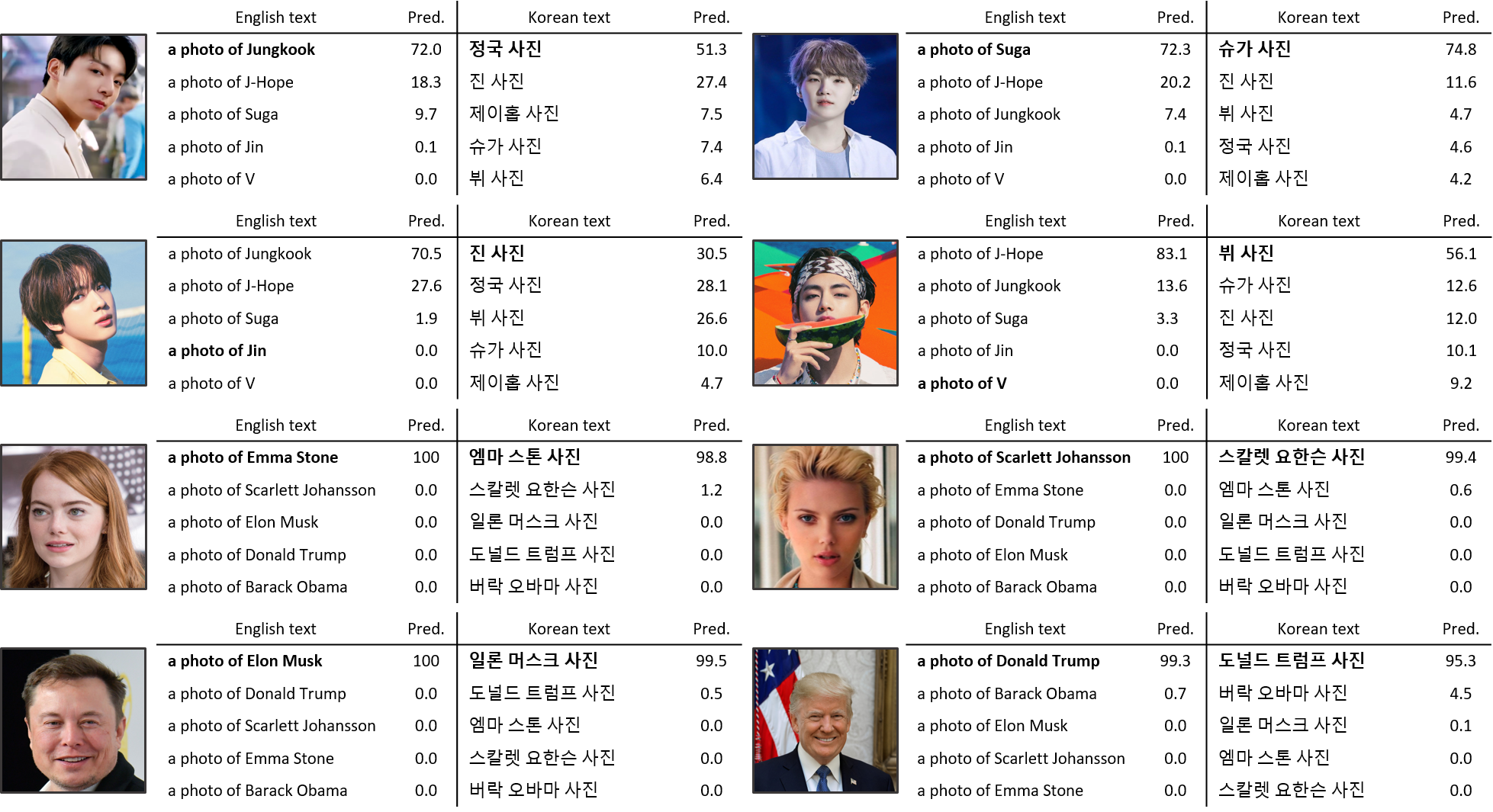}
\caption{\textbf{Qualitative results of face recognition}. A zero-shot KELIP classifier classifies each celebrity's image with five English and Korean texts. Bold text indicates ground truth and prediction (Pred.) value is in \%.}
\label{fig:face}
\end{figure}

\paragraph{Qualitative Analysis of Face Recognition}
As we included celebrities' image and text pairs in the training data, we conduct a qualitative analysis of face recognition in Figure~\ref{fig:face}.

%% file: table/tab_hyperparam.tex
\begin{table}[t]
    \begin{subtable}[h]{0.5\textwidth}
        \begin{adjustbox}{width=0.8\textwidth,center}
        \centering
        \begin{tabular}{lcc}
        \toprule
        Config                 & KELIP$_{\text{yfcc}}$       & KELIP             \\
        \midrule
        Optimizer              & \multicolumn{2}{c}{AdamW}             \\
        Learning rate          & \multicolumn{2}{c}{1.5e-3}          \\
        Weight decay           & \multicolumn{2}{c}{0.5}          \\
        Batch size             & \multicolumn{2}{c}{40960}             \\
        Learning rate schedule & \multicolumn{2}{c}{cosine decay}      \\
        Warmup steps           & \multicolumn{2}{c}{1000}              \\
        Augmentation           & \multicolumn{2}{c}{RandomResizedCrop} \\
        Epochs                 & 50           & 3                 \\
        Mask ratio             & \multicolumn{2}{c}{0.75}              \\
        Decoder layers         & \multicolumn{2}{c}{4}                \\
        \bottomrule
        \end{tabular}
        \end{adjustbox}
    \caption{MAE pre-training setting}
    \label{table:pre-training}
    \end{subtable}
\hfill
    \begin{subtable}[h]{0.5\columnwidth}
        \begin{adjustbox}{width=0.8\columnwidth,center}
        \centering
        \begin{tabular}{lcc}
        \toprule
        Config                 & KELIP$_{\text{yfcc}}$         & KELIP           \\
        \midrule
        Backbone               & \multicolumn{2}{c}{ViT-B/32}          \\
        Optimizer              & \multicolumn{2}{c}{AdamW}             \\
        Learning rate          & 5e-4            & 1.2e-3        \\
        Weight decay           & \multicolumn{2}{c}{0.2}               \\
        Batch size             & 4080                & 65600           \\
        Learning rate schedule & \multicolumn{2}{c}{cosine dcay}       \\
        Warmup steps           & 3500                & 2000            \\
        Augmentation           & \multicolumn{2}{c}{RandomResizedCrop} \\
        Epochs                 & 25                  & 32              \\
        Number of multi-crop   & -                   & 1              \\
        \bottomrule
        \end{tabular}
        \end{adjustbox}
    \caption{KELIP fine-tuning setting}
    \label{table:fine-tuning}
    \end{subtable}
\caption{\textbf{Hyper-parameters of MAE pre-training and KELIP fine-tuning.} KELIP$_{\text{yfcc}}$ is a model trained with small dataset (\ie YFCC15M), while KELIP is a model trained with large dataset (\ie 1.1 billion Korean and English).}
\label{table:hyperparam}
\end{table}

%% file: main_arxiv.bbl
\begin{thebibliography}{71}
\providecommand{\natexlab}[1]{#1}
\providecommand{\url}[1]{\texttt{#1}}
\expandafter\ifx\csname urlstyle\endcsname\relax
  \providecommand{\doi}[1]{doi: #1}\else
  \providecommand{\doi}{doi: \begingroup \urlstyle{rm}\Url}\fi

\bibitem[Ben-Younes et~al.(2017)Ben-Younes, Cadene, Cord, and
  Thome]{ben2017mutan}
Hedi Ben-Younes, R{\'e}mi Cadene, Matthieu Cord, and Nicolas Thome.
\newblock Mutan: Multimodal tucker fusion for visual question answering.
\newblock In \emph{Proc. {ICCV}}, 2017.

\bibitem[Bossard et~al.(2014)Bossard, Guillaumin, and
  Van~Gool]{bossard2014food}
Lukas Bossard, Matthieu Guillaumin, and Luc Van~Gool.
\newblock Food-101--mining discriminative components with random forests.
\newblock In \emph{Proc. {ECCV}}, 2014.

\bibitem[Brown et~al.(2020)Brown, Mann, Ryder, Subbiah, Kaplan, Dhariwal,
  Neelakantan, Shyam, Sastry, Askell, et~al.]{brown2020language}
Tom Brown, Benjamin Mann, Nick Ryder, Melanie Subbiah, Jared~D Kaplan, Prafulla
  Dhariwal, Arvind Neelakantan, Pranav Shyam, Girish Sastry, Amanda Askell,
  et~al.
\newblock Language models are few-shot learners.
\newblock \emph{Proc. {NeurIPS}}, 2020.

\bibitem[Caron et~al.(2020)Caron, Misra, Mairal, Goyal, Bojanowski, and
  Joulin]{caron2020swav}
Mathilde Caron, Ishan Misra, Julien Mairal, Priya Goyal, Piotr Bojanowski, and
  Armand Joulin.
\newblock Unsupervised learning of visual features by contrasting cluster
  assignments.
\newblock In \emph{Proc. {NeurIPS}}, 2020.

\bibitem[Caron et~al.(2021)Caron, Touvron, Misra, J\'egou, Mairal, Bojanowski,
  and Joulin]{caron2021dino}
Mathilde Caron, Hugo Touvron, Ishan Misra, Herv\'e J\'egou, Julien Mairal,
  Piotr Bojanowski, and Armand Joulin.
\newblock Emerging properties in self-supervised vision transformers.
\newblock \emph{Proc. {ICCV}}, 2021.

\bibitem[Chen et~al.(2020{\natexlab{a}})Chen, Kornblith, Norouzi, and
  Hinton]{chen2020simclr}
Ting Chen, Simon Kornblith, Mohammad Norouzi, and Geoffrey Hinton.
\newblock A simple framework for contrastive learning of visual
  representations.
\newblock In \emph{Proc. {ICML}}, 2020{\natexlab{a}}.

\bibitem[Chen et~al.(2020{\natexlab{b}})Chen, Li, Yu, El~Kholy, Ahmed, Gan,
  Cheng, and Liu]{chen2020uniter}
Yen-Chun Chen, Linjie Li, Licheng Yu, Ahmed El~Kholy, Faisal Ahmed, Zhe Gan,
  Yu~Cheng, and Jingjing Liu.
\newblock Uniter: Universal image-text representation learning.
\newblock In \emph{Proc. {ECCV}}, 2020{\natexlab{b}}.

\bibitem[Cheng et~al.(2017)Cheng, Han, and Lu]{cheng2017remote}
Gong Cheng, Junwei Han, and Xiaoqiang Lu.
\newblock Remote sensing image scene classification: Benchmark and state of the
  art.
\newblock \emph{Proceedings of the IEEE}, 2017.

\bibitem[Cimpoi et~al.(2014)Cimpoi, Maji, Kokkinos, Mohamed, , and
  Vedaldi]{cimpoi14describing}
M.~Cimpoi, S.~Maji, I.~Kokkinos, S.~Mohamed, , and A.~Vedaldi.
\newblock Describing textures in the wild.
\newblock In \emph{Proc. {CVPR}}, 2014.

\bibitem[Coates et~al.(2011)Coates, Ng, and Lee]{coates2011analysis}
Adam Coates, Andrew Ng, and Honglak Lee.
\newblock An analysis of single-layer networks in unsupervised feature
  learning.
\newblock In \emph{Proc. {AISTATS}}, 2011.

\bibitem[Deng et~al.(2009)Deng, Dong, Socher, Li, Li, and
  Fei-Fei]{deng2009imagenet}
Jia Deng, Wei Dong, Richard Socher, Li-Jia Li, Kai Li, and Li~Fei-Fei.
\newblock Imagenet: A large-scale hierarchical image database.
\newblock In \emph{Proc. {CVPR}}, 2009.

\bibitem[Devlin et~al.(2019)Devlin, Chang, Lee, and Toutanova]{devlin2019bert}
Jacob Devlin, Ming-Wei Chang, Kenton Lee, and Kristina Toutanova.
\newblock Bert: Pre-training of deep bidirectional transformers for language
  understanding.
\newblock In \emph{{ACL}}, 2019.

\bibitem[Dhariwal \& Nichol(2021)Dhariwal and Nichol]{dhariwal2021diffusion}
Prafulla Dhariwal and Alexander Nichol.
\newblock Diffusion models beat gans on image synthesis.
\newblock \emph{Proc. {NeurIPS}}, 2021.

\bibitem[Dosovitskiy et~al.(2021)Dosovitskiy, Beyer, Kolesnikov, Weissenborn,
  Zhai, Unterthiner, Dehghani, Minderer, Heigold, Gelly, Uszkoreit, and
  Houlsby]{dosovitskiy2021an}
Alexey Dosovitskiy, Lucas Beyer, Alexander Kolesnikov, Dirk Weissenborn,
  Xiaohua Zhai, Thomas Unterthiner, Mostafa Dehghani, Matthias Minderer, Georg
  Heigold, Sylvain Gelly, Jakob Uszkoreit, and Neil Houlsby.
\newblock An image is worth 16x16 words: Transformers for image recognition at
  scale.
\newblock In \emph{Proc. {ICLR}}, 2021.

\bibitem[Ethayarajh et~al.(2018)Ethayarajh, Duvenaud, and
  Hirst]{ethayarajh2018towards}
Kawin Ethayarajh, David Duvenaud, and Graeme Hirst.
\newblock Towards understanding linear word analogies.
\newblock \emph{arXiv preprint arXiv:1810.04882}, 2018.

\bibitem[Fabbrizzi et~al.(2021)Fabbrizzi, Papadopoulos, Ntoutsi, and
  Kompatsiaris]{fabbrizzi2021survey}
Simone Fabbrizzi, Symeon Papadopoulos, Eirini Ntoutsi, and Ioannis
  Kompatsiaris.
\newblock A survey on bias in visual datasets.
\newblock \emph{arXiv preprint arXiv:2107.07919}, 2021.

\bibitem[Fukui et~al.(2016)Fukui, Park, Yang, Rohrbach, Darrell, and
  Rohrbach]{fukui2016multimodal}
Akira Fukui, Dong~Huk Park, Daylen Yang, Anna Rohrbach, Trevor Darrell, and
  Marcus Rohrbach.
\newblock Multimodal compact bilinear pooling for visual question answering and
  visual grounding.
\newblock \emph{arXiv preprint arXiv:1606.01847}, 2016.

\bibitem[Galatolo et~al.(2021)Galatolo, Cimino, and
  Vaglini]{galatolo2021generating}
Federico~A Galatolo, Mario~GCA Cimino, and Gigliola Vaglini.
\newblock Generating images from caption and vice versa via clip-guided
  generative latent space search.
\newblock \emph{arXiv preprint arXiv:2102.01645}, 2021.

\bibitem[Goodfellow et~al.(2013)Goodfellow, Erhan, Carrier, Courville, Mirza,
  Hamner, Cukierski, Tang, Thaler, Lee, et~al.]{goodfellow2013challenges}
Ian~J Goodfellow, Dumitru Erhan, Pierre~Luc Carrier, Aaron Courville, Mehdi
  Mirza, Ben Hamner, Will Cukierski, Yichuan Tang, David Thaler, Dong-Hyun Lee,
  et~al.
\newblock Challenges in representation learning: A report on three machine
  learning contests.
\newblock In \emph{Proc. {NeurIPS}}, 2013.

\bibitem[Grill et~al.(2020)Grill, Strub, Altch{\'e}, Tallec, Richemond,
  Buchatskaya, Doersch, Avila~Pires, Guo, Gheshlaghi~Azar,
  et~al.]{grill2020bootstrap}
Jean-Bastien Grill, Florian Strub, Florent Altch{\'e}, Corentin Tallec, Pierre
  Richemond, Elena Buchatskaya, Carl Doersch, Bernardo Avila~Pires, Zhaohan
  Guo, Mohammad Gheshlaghi~Azar, et~al.
\newblock Bootstrap your own latent-a new approach to self-supervised learning.
\newblock \emph{Proc. {NeurIPS}}, 2020.

\bibitem[He et~al.(2020)He, Fan, Wu, Xie, and Girshick]{he2020moco}
Kaiming He, Haoqi Fan, Yuxin Wu, Saining Xie, and Ross Girshick.
\newblock Momentum contrast for unsupervised visual representation learning.
\newblock In \emph{Proc. {CVPR}}, 2020.

\bibitem[He et~al.(2021)He, Chen, Xie, Li, Doll{\'a}r, and
  Girshick]{he2021masked}
Kaiming He, Xinlei Chen, Saining Xie, Yanghao Li, Piotr Doll{\'a}r, and Ross
  Girshick.
\newblock Masked autoencoders are scalable vision learners.
\newblock \emph{arXiv preprint arXiv:2111.06377}, 2021.

\bibitem[Helber et~al.(2019)Helber, Bischke, Dengel, and
  Borth]{helber2019eurosat}
Patrick Helber, Benjamin Bischke, Andreas Dengel, and Damian Borth.
\newblock Eurosat: A novel dataset and deep learning benchmark for land use and
  land cover classification.
\newblock \emph{IEEE Journal of Selected Topics in Applied Earth Observations
  and Remote Sensing}, 2019.

\bibitem[Ho et~al.(2020)Ho, Jain, and Abbeel]{ho2020denoising}
Jonathan Ho, Ajay Jain, and Pieter Abbeel.
\newblock Denoising diffusion probabilistic models.
\newblock \emph{Proc. {NeurIPS}}, 2020.

\bibitem[Jain et~al.(2021)Jain, Guo, Srinivasan, Chen, Kudugunta, Jia, Yang,
  and Baldridge]{jain2021mural}
Aashi Jain, Mandy Guo, Krishna Srinivasan, Ting Chen, Sneha Kudugunta, Chao
  Jia, Yinfei Yang, and Jason Baldridge.
\newblock Mural: multimodal, multitask retrieval across languages.
\newblock \emph{arXiv preprint arXiv:2109.05125}, 2021.

\bibitem[Jia et~al.(2021)Jia, Yang, Xia, Chen, Parekh, Pham, Le, Sung, Li, and
  Duerig]{jia2021scaling}
Chao Jia, Yinfei Yang, Ye~Xia, Yi-Ting Chen, Zarana Parekh, Hieu Pham, Quoc Le,
  Yun-Hsuan Sung, Zhen Li, and Tom Duerig.
\newblock Scaling up visual and vision-language representation learning with
  noisy text supervision.
\newblock In \emph{Proc. {ICML}}, 2021.

\bibitem[Johnson et~al.(2017)Johnson, Hariharan, Van Der~Maaten, Fei-Fei,
  Lawrence~Zitnick, and Girshick]{johnson2017clevr}
Justin Johnson, Bharath Hariharan, Laurens Van Der~Maaten, Li~Fei-Fei,
  C~Lawrence~Zitnick, and Ross Girshick.
\newblock Clevr: A diagnostic dataset for compositional language and elementary
  visual reasoning.
\newblock In \emph{Proc. {CVPR}}, 2017.

\bibitem[Krause et~al.(2013)Krause, Stark, Deng, and Fei-Fei]{krause20133d}
Jonathan Krause, Michael Stark, Jia Deng, and Li~Fei-Fei.
\newblock 3d object representations for fine-grained categorization.
\newblock In \emph{Proc. {ICCV-W}}, 2013.

\bibitem[Krizhevsky \& Hinton(2009)Krizhevsky and
  Hinton]{krizhevsky2009learning}
Alex Krizhevsky and Geoffrey Hinton.
\newblock Learning multiple layers of features from tiny images.
\newblock Technical report, {UT}, 2009.

\bibitem[LeCun et~al.(1998)LeCun, Bottou, Bengio, and Haffner]{lecun1998mnist}
Yann LeCun, L{\'e}on Bottou, Yoshua Bengio, and Patrick Haffner.
\newblock Gradient-based learning applied to document recognition.
\newblock \emph{Proceedings of the IEEE}, 1998.

\bibitem[Li et~al.(2017)Li, Jabri, Joulin, and Van Der~Maaten]{li2017learning}
Ang Li, Allan Jabri, Armand Joulin, and Laurens Van Der~Maaten.
\newblock Learning visual n-grams from web data.
\newblock In \emph{Proc. {ICCV}}, 2017.

\bibitem[Li et~al.(2020)Li, Duan, Fang, Gong, and Jiang]{li2020unicoder}
Gen Li, Nan Duan, Yuejian Fang, Ming Gong, and Daxin Jiang.
\newblock Unicoder-vl: A universal encoder for vision and language by
  cross-modal pre-training.
\newblock In \emph{Proc. {AAAI}}, 2020.

\bibitem[Li et~al.(2022)Li, Li, Xiong, and Hoi]{li2022blip}
Junnan Li, Dongxu Li, Caiming Xiong, and Steven Hoi.
\newblock Blip: Bootstrapping language-image pre-training for unified
  vision-language understanding and generation.
\newblock \emph{arXiv preprint arXiv:2201.12086}, 2022.

\bibitem[Li et~al.(2019)Li, Yatskar, Yin, Hsieh, and Chang]{li2019visualbert}
Liunian~Harold Li, Mark Yatskar, Da~Yin, Cho-Jui Hsieh, and Kai-Wei Chang.
\newblock Visualbert: A simple and performant baseline for vision and language.
\newblock \emph{arXiv preprint arXiv:1908.03557}, 2019.

\bibitem[Li et~al.(2021)Li, Liang, Zhao, Cui, Ouyang, Shao, Yu, and
  Yan]{li2021supervision}
Yangguang Li, Feng Liang, Lichen Zhao, Yufeng Cui, Wanli Ouyang, Jing Shao,
  Fengwei Yu, and Junjie Yan.
\newblock Supervision exists everywhere: A data efficient contrastive
  language-image pre-training paradigm.
\newblock \emph{arXiv preprint arXiv:2110.05208}, 2021.

\bibitem[Lin et~al.(2014)Lin, Maire, Belongie, Hays, Perona, Ramanan,
  Doll{\'a}r, and Zitnick]{lin2014microsoft}
Tsung-Yi Lin, Michael Maire, Serge Belongie, James Hays, Pietro Perona, Deva
  Ramanan, Piotr Doll{\'a}r, and C~Lawrence Zitnick.
\newblock Microsoft coco: Common objects in context.
\newblock In \emph{Proc. {ECCV}}, 2014.

\bibitem[Lu et~al.(2019)Lu, Batra, Parikh, and Lee]{lu2019vilbert}
Jiasen Lu, Dhruv Batra, Devi Parikh, and Stefan Lee.
\newblock Vilbert: Pretraining task-agnostic visiolinguistic representations
  for vision-and-language tasks.
\newblock \emph{Proc. {NeurIPS}}, 2019.

\bibitem[Luo et~al.(2021)Luo, Ji, Zhong, Chen, Lei, Duan, and
  Li]{luo2021clip4clip}
Huaishao Luo, Lei Ji, Ming Zhong, Yang Chen, Wen Lei, Nan Duan, and Tianrui Li.
\newblock Clip4clip: An empirical study of clip for end to end video clip
  retrieval.
\newblock \emph{arXiv preprint arXiv:2104.08860}, 2021.

\bibitem[Mehrabi et~al.(2021)Mehrabi, Morstatter, Saxena, Lerman, and
  Galstyan]{mehrabi2021survey}
Ninareh Mehrabi, Fred Morstatter, Nripsuta Saxena, Kristina Lerman, and Aram
  Galstyan.
\newblock A survey on bias and fairness in machine learning.
\newblock \emph{ACM Computing Surveys (CSUR)}, 2021.

\bibitem[Micikevicius et~al.(2017)Micikevicius, Narang, Alben, Diamos, Elsen,
  Garcia, Ginsburg, Houston, Kuchaiev, Venkatesh,
  et~al.]{micikevicius2017mixed}
Paulius Micikevicius, Sharan Narang, Jonah Alben, Gregory Diamos, Erich Elsen,
  David Garcia, Boris Ginsburg, Michael Houston, Oleksii Kuchaiev, Ganesh
  Venkatesh, et~al.
\newblock Mixed precision training.
\newblock \emph{arXiv preprint arXiv:1710.03740}, 2017.

\bibitem[Misra \& van~der Maaten(2020)Misra and van~der
  Maaten]{misra2020pretext}
Ishan Misra and Laurens van~der Maaten.
\newblock Self-supervised learning of pretext-invariant representations.
\newblock \emph{Proc. {CVPR}}, 2020.

\bibitem[Mokady et~al.(2021)Mokady, Hertz, and Bermano]{mokady2021clipcap}
Ron Mokady, Amir Hertz, and Amit~H Bermano.
\newblock Clipcap: Clip prefix for image captioning.
\newblock \emph{arXiv preprint arXiv:2111.09734}, 2021.

\bibitem[Mu et~al.(2021)Mu, Kirillov, Wagner, and Xie]{mu2021slip}
Norman Mu, Alexander Kirillov, David Wagner, and Saining Xie.
\newblock Slip: Self-supervision meets language-image pre-training.
\newblock \emph{arXiv preprint arXiv:2112.12750}, 2021.

\bibitem[Oord et~al.(2018)Oord, Li, and Vinyals]{oord2018cpc}
Aaron van~den Oord, Yazhe Li, and Oriol Vinyals.
\newblock Representation learning with contrastive predictive coding.
\newblock \emph{arXiv:1807.03748}, 2018.

\bibitem[Paszke et~al.(2019)Paszke, Gross, Massa, Lerer, Bradbury, Chanan,
  Killeen, Lin, Gimelshein, Antiga, Desmaison, Kopf, Yang, DeVito, Raison,
  Tejani, Chilamkurthy, Steiner, Fang, Bai, and Chintala]{pytorch}
Adam Paszke, Sam Gross, Francisco Massa, Adam Lerer, James Bradbury, Gregory
  Chanan, Trevor Killeen, Zeming Lin, Natalia Gimelshein, Luca Antiga, Alban
  Desmaison, Andreas Kopf, Edward Yang, Zachary DeVito, Martin Raison, Alykhan
  Tejani, Sasank Chilamkurthy, Benoit Steiner, Lu~Fang, Junjie Bai, and Soumith
  Chintala.
\newblock {PyTorch}: {A}n imperative style, high-performance deep learning
  library.
\newblock In \emph{Proc. {NeurIPS}}, 2019.

\bibitem[Patashnik et~al.(2021)Patashnik, Wu, Shechtman, Cohen-Or, and
  Lischinski]{patashnik2021styleclip}
Or~Patashnik, Zongze Wu, Eli Shechtman, Daniel Cohen-Or, and Dani Lischinski.
\newblock Styleclip: Text-driven manipulation of stylegan imagery.
\newblock In \emph{Proceedings of the IEEE/CVF International Conference on
  Computer Vision}, pp.\  2085--2094, 2021.

\bibitem[Pham et~al.(2021)Pham, Dai, Ghiasi, Liu, Yu, Luong, Tan, and
  Le]{pham2021combined}
Hieu Pham, Zihang Dai, Golnaz Ghiasi, Hanxiao Liu, Adams~Wei Yu, Minh-Thang
  Luong, Mingxing Tan, and Quoc~V Le.
\newblock Combined scaling for zero-shot transfer learning.
\newblock \emph{arXiv preprint arXiv:2111.10050}, 2021.

\bibitem[Plummer et~al.(2015)Plummer, Wang, Cervantes, Caicedo, Hockenmaier,
  and Lazebnik]{plummer2015flickr30k}
Bryan~A Plummer, Liwei Wang, Chris~M Cervantes, Juan~C Caicedo, Julia
  Hockenmaier, and Svetlana Lazebnik.
\newblock Flickr30k entities: Collecting region-to-phrase correspondences for
  richer image-to-sentence models.
\newblock In \emph{Proc. {ICCV}}, 2015.

\bibitem[Qi et~al.(2020)Qi, Su, Song, Cui, Bharti, and
  Sacheti]{qi2020imagebert}
Di~Qi, Lin Su, Jia Song, Edward Cui, Taroon Bharti, and Arun Sacheti.
\newblock Imagebert: Cross-modal pre-training with large-scale weak-supervised
  image-text data.
\newblock \emph{arXiv preprint arXiv:2001.07966}, 2020.

\bibitem[Radford et~al.(2018)Radford, Narasimhan, Salimans, and
  Sutskever]{radford2018improving}
Alec Radford, Karthik Narasimhan, Tim Salimans, and Ilya Sutskever.
\newblock Improving language understanding by generative pre-training (2018),
  2018.

\bibitem[Radford et~al.(2019)Radford, Wu, Child, Luan, Amodei, Sutskever,
  et~al.]{radford2019language}
Alec Radford, Jeffrey Wu, Rewon Child, David Luan, Dario Amodei, Ilya
  Sutskever, et~al.
\newblock Language models are unsupervised multitask learners.
\newblock \emph{OpenAI blog}, 2019.

\bibitem[Radford et~al.(2021{\natexlab{a}})Radford, Kim, Hallacy, Ramesh, Goh,
  Agarwal, Sastry, Askell, Mishkin, Clark, Krueger, and
  Sutskever]{radford2021learning}
Alec Radford, Jong~Wook Kim, Chris Hallacy, Aditya Ramesh, Gabriel Goh,
  Sandhini Agarwal, Girish Sastry, Amanda Askell, Pamela Mishkin, Jack Clark,
  Gretchen Krueger, and Ilya Sutskever.
\newblock Learning transferable visual models from natural language
  supervision.
\newblock In \emph{Proc. {ICML}}, 2021{\natexlab{a}}.

\bibitem[Radford et~al.(2021{\natexlab{b}})Radford, Kim, Hallacy, Ramesh, Goh,
  Agarwal, Sastry, Askell, Mishkin, Clark, et~al.]{radford2021clip}
Alec Radford, Jong~Wook Kim, Chris Hallacy, Aditya Ramesh, Gabriel Goh,
  Sandhini Agarwal, Girish Sastry, Amanda Askell, Pamela Mishkin, Jack Clark,
  et~al.
\newblock Learning transferable visual models from natural language
  supervision.
\newblock In \emph{Proc. {ICML}}, 2021{\natexlab{b}}.

\bibitem[Ramesh et~al.(2021)Ramesh, Pavlov, Goh, Gray, Voss, Radford, Chen, and
  Sutskever]{ramesh2021zero}
Aditya Ramesh, Mikhail Pavlov, Gabriel Goh, Scott Gray, Chelsea Voss, Alec
  Radford, Mark Chen, and Ilya Sutskever.
\newblock Zero-shot text-to-image generation.
\newblock In \emph{Proc. {ICML}}, 2021.

\bibitem[Real et~al.(2017)Real, Shlens, Mazzocchi, Pan, and
  Vanhoucke]{real2017youtube}
Esteban Real, Jonathon Shlens, Stefano Mazzocchi, Xin Pan, and Vincent
  Vanhoucke.
\newblock Youtube-boundingboxes: A large high-precision human-annotated data
  set for object detection in video.
\newblock In \emph{Proc. {CVPR}}, 2017.

\bibitem[Schuhmann et~al.(2021)Schuhmann, Vencu, Beaumont, Kaczmarczyk, Mullis,
  Katta, Coombes, Jitsev, and Komatsuzaki]{schuhmann2021laion}
Christoph Schuhmann, Richard Vencu, Romain Beaumont, Robert Kaczmarczyk,
  Clayton Mullis, Aarush Katta, Theo Coombes, Jenia Jitsev, and Aran
  Komatsuzaki.
\newblock Laion-400m: Open dataset of clip-filtered 400 million image-text
  pairs.
\newblock \emph{arXiv preprint arXiv:2111.02114}, 2021.

\bibitem[Sharma et~al.(2018)Sharma, Ding, Goodman, and
  Soricut]{sharma2018conceptual}
Piyush Sharma, Nan Ding, Sebastian Goodman, and Radu Soricut.
\newblock Conceptual captions: A cleaned, hypernymed, image alt-text dataset
  for automatic image captioning.
\newblock In \emph{Proceedings of the 56th Annual Meeting of the Association
  for Computational Linguistics (Volume 1: Long Papers)}, 2018.

\bibitem[Shin et~al.(2021)Shin, Cho, Ko, and Gu]{shin2021rtic}
Minchul Shin, Yoonjae Cho, Byungsoo Ko, and Geonmo Gu.
\newblock Rtic: Residual learning for text and image composition using graph
  convolutional network.
\newblock \emph{arXiv preprint arXiv:2104.03015}, 2021.

\bibitem[Shonenkov et~al.(2022)Shonenkov, Kuznetsov, Dimitrov, Shavrina,
  Chesakov, Maltseva, Fenogenova, Pavlov, Emelyanov, Markov,
  et~al.]{shonenkov2022ruclip}
Alex Shonenkov, Andrey Kuznetsov, Denis Dimitrov, Tatyana Shavrina, Daniil
  Chesakov, Anastasia Maltseva, Alena Fenogenova, Igor Pavlov, Anton Emelyanov,
  Sergey Markov, et~al.
\newblock Ruclip--new models and experiments: a technical report.
\newblock \emph{arXiv preprint arXiv:2202.10784}, 2022.

\bibitem[Srinivasan et~al.(2021)Srinivasan, Raman, Chen, Bendersky, and
  Najork]{srinivasan2021wit}
Krishna Srinivasan, Karthik Raman, Jiecao Chen, Michael Bendersky, and Marc
  Najork.
\newblock Wit: Wikipedia-based image text dataset for multimodal multilingual
  machine learning.
\newblock \emph{arXiv preprint arXiv:2103.01913}, 2021.

\bibitem[Stallkamp et~al.(2012)Stallkamp, Schlipsing, Salmen, and
  Igel]{stallkamp2012man}
Johannes Stallkamp, Marc Schlipsing, Jan Salmen, and Christian Igel.
\newblock Man vs. computer: Benchmarking machine learning algorithms for
  traffic sign recognition.
\newblock \emph{Neural networks}, 2012.

\bibitem[Thomee et~al.(2016)Thomee, Shamma, Friedland, Elizalde, Ni, Poland,
  Borth, and Li]{thomee2016yfcc100m}
Bart Thomee, David~A Shamma, Gerald Friedland, Benjamin Elizalde, Karl Ni,
  Douglas Poland, Damian Borth, and Li-Jia Li.
\newblock Yfcc100m: The new data in multimedia research.
\newblock \emph{Communications of the ACM}, 2016.

\bibitem[Vaswani et~al.(2017)Vaswani, Shazeer, Parmar, Uszkoreit, Jones, Gomez,
  Kaiser, and Polosukhin]{Vaswani2017AttentionIA}
Ashish Vaswani, Noam Shazeer, Niki Parmar, Jakob Uszkoreit, Llion Jones,
  Aidan~N. Gomez, Lukasz Kaiser, and Illia Polosukhin.
\newblock Attention is all you need.
\newblock In \emph{Proc. {NeurIPS}}, 2017.

\bibitem[Wah et~al.(2011)Wah, Branson, Welinder, Perona, and
  Belongie]{wah2011cub}
C.~Wah, S.~Branson, P.~Welinder, P.~Perona, and S.~Belongie.
\newblock {The Caltech-UCSD Birds-200-2011 Dataset}.
\newblock Technical report, California Institute of Technology, 2011.

\bibitem[Wu et~al.(2018)Wu, Xiong, Yu, and Lin]{wu2018unsupervised}
Zhirong Wu, Yuanjun Xiong, Stella~X Yu, and Dahua Lin.
\newblock Unsupervised feature learning via non-parametric instance
  discrimination.
\newblock In \emph{Proc. {CVPR}}, 2018.

\bibitem[Xiao et~al.(2010)Xiao, Hays, Ehinger, Oliva, and
  Torralba]{xiao2010sun}
Jianxiong Xiao, James Hays, Krista~A Ehinger, Aude Oliva, and Antonio Torralba.
\newblock Sun database: Large-scale scene recognition from abbey to zoo.
\newblock In \emph{Proc. {CVPR}}, 2010.

\bibitem[Yao et~al.(2021)Yao, Huang, Hou, Lu, Niu, Xu, Liang, Li, Jiang, and
  Xu]{yao2021filip}
Lewei Yao, Runhui Huang, Lu~Hou, Guansong Lu, Minzhe Niu, Hang Xu, Xiaodan
  Liang, Zhenguo Li, Xin Jiang, and Chunjing Xu.
\newblock Filip: Fine-grained interactive language-image pre-training.
\newblock \emph{arXiv preprint arXiv:2111.07783}, 2021.

\bibitem[Yu et~al.(2017)Yu, Yu, Fan, and Tao]{yu2017multi}
Zhou Yu, Jun Yu, Jianping Fan, and Dacheng Tao.
\newblock Multi-modal factorized bilinear pooling with co-attention learning
  for visual question answering.
\newblock In \emph{Proc. {ICCV}}, 2017.

\bibitem[Zhai et~al.(2021)Zhai, Wang, Mustafa, Steiner, Keysers, Kolesnikov,
  and Beyer]{zhai2021lit}
Xiaohua Zhai, Xiao Wang, Basil Mustafa, Andreas Steiner, Daniel Keysers,
  Alexander Kolesnikov, and Lucas Beyer.
\newblock Lit: Zero-shot transfer with locked-image text tuning.
\newblock \emph{arXiv preprint arXiv:2111.07991}, 2021.

\bibitem[Zhou et~al.(2021)Zhou, Wei, Wang, Shen, Xie, Yuille, and
  Kong]{zhou2021ibot}
Jinghao Zhou, Chen Wei, Huiyu Wang, Wei Shen, Cihang Xie, Alan Yuille, and Tao
  Kong.
\newblock ibot: Image bert pre-training with online tokenizer.
\newblock \emph{arXiv preprint arXiv:2111.07832}, 2021.

\bibitem[Zhou et~al.(2022)Zhou, Girdhar, Joulin, Kr{\"a}henb{\"u}hl, and
  Misra]{zhou2022detecting}
Xingyi Zhou, Rohit Girdhar, Armand Joulin, Phillip Kr{\"a}henb{\"u}hl, and
  Ishan Misra.
\newblock Detecting twenty-thousand classes using image-level supervision.
\newblock \emph{arXiv preprint arXiv:2201.02605}, 2022.

\end{thebibliography}
